\newcommand{\comment}[1]{}
\documentclass[11pt,letterpaper]{article}
\usepackage{emnlp2016}
\usepackage{times}
\usepackage{latexsym}
\usepackage{url}
\usepackage{color}
\usepackage{float}
\usepackage{tabularx}
\usepackage{subfig}
\usepackage{multirow}
\usepackage{latexsym}
\usepackage{amsmath}
\usepackage{amssymb}
\usepackage{multirow}

\usepackage{pgfplots}
\DeclareMathOperator*{\softmax}{softmax}

\emnlpfinalcopy

\def\emnlppaperid{***}

\allowdisplaybreaks

\def\x{\mathbf{x}}

\def\bz{\mathbf{z}}

\def\h{\textbf{h}}

\def\bb{\mathbf{b}}
\def\bc{\mathbf{c}}
\def\bh{\mathbf{h}}

\def\bW{\mathbf{W}}
\def\bU{\mathbf{U}}

\newenvironment{itemize*}%
  {\begin{itemize}%
    \setlength{\itemsep}{1pt}%
    \setlength{\parskip}{1pt}}%
  {\end{itemize}}
  \newenvironment{enumerate*}%
  {\begin{enumerate}%
    \setlength{\itemsep}{1pt}%
    \setlength{\parskip}{1pt}}%
  {\end{enumerate}}

\title{Cached Long Short-Term Memory Neural Networks \\for Document-Level Sentiment Classification}

\author{Jiacheng Xu\footnotemark[2]\quad Danlu Chen\footnotemark[3]\quad Xipeng Qiu\thanks{\quad Corresponding author.} \footnotemark[3] \quad Xuanjing Huang\footnotemark[3] \\
  Software School, Fudan University  \footnotemark[2]\\
  School of Computer Science, Fudan University \footnotemark[3]\\
  Shanghai Key Laboratory of Intelligent Information Processing, Fudan University\footnotemark[2] \footnotemark[3]\\
  825 Zhangheng Road, Shanghai, China\footnotemark[2] \footnotemark[3]\\
  {\tt \{jcxu13,dlchen13,xpqiu,xjhuang\}@fudan.edu.cn}\\
}

\date{}

\begin{document}

\maketitle

\begin{abstract}
Recently, neural networks have achieved great success on sentiment classification due to their ability to alleviate feature engineering. However, one of the remaining challenges is to model long texts in document-level sentiment classification under a recurrent architecture because of the deficiency of the memory unit.
To address this problem, we present a Cached Long Short-Term Memory neural networks (CLSTM) to capture the overall semantic information in long texts. CLSTM introduces a cache mechanism, which divides memory into several groups with different forgetting rates and thus enables the network to keep sentiment information better within a recurrent unit. The proposed CLSTM outperforms the state-of-the-art models on three publicly available document-level sentiment analysis datasets.

\end{abstract}

\section{Introduction}
Sentiment classification is one of the most widely used natural language processing techniques in many areas, such as E-commerce websites, online social networks, political orientation analyses \cite{wilson2009recognizing,OConnor:2010tn}, etc.
%Generally, we are mostly concerned on sentiment polarity analysis, since it provides a satisfactory knowledge about a client or community?s general opinion.
%It's straightforward to do such an analysis by turning it into a multi-class classification problem, and we can build a classifier as in many other text classification tasks.
%The state-of-the-art methods of sentiment classification are based on various machine learning algorithms with handcrafted features (usually with the help of a sentiment lexicon) \cite{pang2002thumbs,blitzer2007biographies,Tang:2014wc,Diao:2014gn}.

Recently, deep learning approaches \cite{socher2013recursive,kim2014convolutional,chen2015sentence,liu2016recurrent} have gained encouraging results on sentiment classification, which frees researchers from handcrafted feature engineering.
%These models, such as recurrent neural networks (RNNs) \cite{Tang:2015ts}, recursive neural networks (RecNNs) \cite{socher2013recursive}, convolutional neural networks (CNNs) \cite{kalchbrenner2014convolutional,kim2014convolutional}, take vector representation of words in a text as input, and generate a fixed length vector as the representation of the whole text.
Among these methods, Recurrent Neural Networks (RNNs) are one of the most prevalent architectures because of the ability to handle variable-length texts.
% Among these methods, Recurrent Neural Networks (RNNs) are one of the most popular architectures because its model parameters is independent of the length of input sequence, and thus it is desirable to process variable-length texts.
%\danlu{Should clarify here why we choose LSTM rather than other machine learning approach? }

Sentence- or paragraph-level sentiment analysis expects the  model to extract features from limited source of information, while document-level sentiment analysis demands more on selecting and storing global sentiment message from long  texts with noises and redundant local pattern. Simple RNNs are  not powerful enough to handle the overflow and to pick up key sentiment messages from relatively far time-steps .

% RNN and its variants work well in sentence-level or paragraph-level sentiment analysis for their ability to extract and ``remember'' features with their memory units, but they are inadequate to capture the overall semantic information when the text is too long, because the redundant noise is increasing and RNNs are likely to forget key sentiment message from relatively far time-steps \cite{karpathy2015visualizing}.

%which leads to a decline of accuracy on sentiment analysis.
Efforts have been made to solve such a scalability problem on long texts by extracting semantic information hierarchically \cite{Tang:2015ts,Tai:2015wp}, which first obtain sentence representations
% by convolution layers or non-neural methods
and then combine them to generate  high-level document embeddings. However, some of these solutions either rely on explicit \textit{a priori} structural assumptions or discard the order information within a sentence, which are vulnerable to sudden change or twists in texts especially a long-range one \cite{McDonald:2007uh,Mikolov:2013wc}. Recurrent models match people's intuition of reading word by word and are capable to model the intrinsic relations between sentences. By keeping the word order, RNNs could extract the sentence representation implicitly and meanwhile analyze the semantic meaning of a whole document without any explicit boundary. %which may not utilize RNN's goodness on intrinsic semantic information extraction. Moreover, these models .
%which avoid the scalability problem by shortening the length of the sequence before feeding it to a recurrent model.
%Such a strategy would inevitably incur systematic bias and do not squeeze the neural network's potential to capture overall emotional message and thus %\danlu{this paragraph needs to be rewritten, or move to related work} %When a person is reading, \danlu{he or she extracts multi-scale information from words, sentences and paragraphs}, which enlightens us on a new architecture of recurrent neural network.

%Although these methods alleviates the long-range dependence problem, document-level sentiment classification remains challenging.

Partially inspired by neural structure of human brain and computer system architecture, we present the Cached Long Short-Term Memory neural networks (CLSTM) to capture the long-range sentiment information. In the dual store memory model proposed by \newcite{atkinson1968human}, memories can reside in the short-term ``buffer'' for a limited time while they are simultaneously strengthening their associations in long-term memory. Accordingly, CLSTM equips a standard LSTM with a similar cache mechanism, whose internal memory is divided into several groups with different forgetting rates. A group with high forgetting rate plays a role as a cache in our model, bridging and transiting the information to groups with relatively lower forgetting rates. With different forgetting rates, CLSTM learns to capture, remember and forget semantics information through a very long distance.

% and thus enable the RNN-like models to directly analyze opinion in a text, of which the length is scaling up.

   % using a uniform CIFG gate with different forgetting rates, which is simplified and merged from the so-called input gate and output gate.

% \cite{Chung:2015vxa} and \cite{Greff:2015wv},

%\danlu{actually, I think the sensory adaptation theory is much more suitable for ``inspiration'' purpose. The LSTM is not incapable to remember message via a long range. Rather, it remember too much at the beginning of trainging stage and somehow is sick as Hyperthymesia, and in theory LSTM should learn how to remember things selectively (i.e. the acquisition of sensory adaptation) given a considerably long training time. Our method is teaching LSTM not to remember everything at the very beginning or it will stuck in a local optimum solution. I am just rambling =w=}

Our main contributions are as follows:
\begin{itemize*}
  \item We introduce a cache mechanism to diversify the internal memory into several distinct groups with different memory cycles by squashing their forgetting rates. As a result, our model can capture the local and global emotional information, thereby better summarizing and analyzing sentiment on long texts in an RNN fashion.
  \item Benefiting from long-term memory unit with a low forgetting rate, we could keep the gradient stable in the long back-propagation process. Hence, our model could converge faster than a standard LSTM.
    \item Our model outperforms state-of-the-art methods by a large margin on three document-level datasets (Yelp 2013, Yelp 2014 and IMDB). It worth noticing that some of the previous methods have utilized extra user and product information.
\end{itemize*}

%Our code and model will be released to the public upon acceptance of the paper.

\section{Related Work}\label{sec:rel}

In this section, we briefly introduce related work in two areas: First, we discuss the existing document-level sentiment classification approaches; Second, we discuss some variants of LSTM which address the problem on storing the long-term information.

\subsection{Document-level Sentiment Classification}
Document-level sentiment classification is a sticky task in sentiment analysis \cite{pang2008opinion}, which is to infer the sentiment polarity or intensity of a whole document. The most challenging part is that not every part of the document is equally informative for inferring the sentiment of the whole document \cite{Pang2004-se,yessenalina2010multi}. Various methods have been investigated and explored over years \cite{wilson2005recognizing,pang2008opinion,pak2010twitter,yessenalina2010multi,moraes2013document}. Most of these methods depend on traditional machine learning algorithms, and are in need of effective handcrafted features.

%The terminology document-level is ambiguous, we use paragraph-level to .
%In other word, plain features such as CBOW or n-gram will bring too much noise into the models.

%Previous non-neural methods mostly concentrated on constructing a suitable hierarchical architecture to model a text,

Recently, neural network based methods are prevalent due to their ability of learning discriminative features from data \cite{socher2013recursive,le2014distributed,Tang:2015ts}.
%\newcite{irsoy2014opinion} using recurrent neural networks to model short texts;
 \newcite{zhu2015long} and \newcite{Tai:2015wp} integrate a tree-structured model into LSTM for better semantic composition; \newcite{bhatia2015better} enhances document-level sentiment analysis by using extra discourse paring results.  Most of these models work well on sentence-level or paragraph-level sentiment classification. When it comes to the document-level sentiment classification, a bottom-up hierarchical strategy is often adopted to alleviate the model complexity \cite{denil2014modelling,Tang:2015vb}.

\subsection{Memory Augmented Recurrent Models}

Although it is widely accepted that LSTM has more long-lasting memory units than RNNs, it still suffers from ``forgetting'' information which is too far away from the current point \cite{le2015simple,karpathy2015visualizing}.  Such a scalability problem of LSTMs is crucial to extend some previous sentence-level work to document-level sentiment analysis.
%In other words, classic LSTM fails to capture necessary number of sentiment features in a long text, especially when there are hundreds of words in it.

Various models have been proposed to increase the ability of LSTMs to store long-range information \cite{le2015simple,salehinejad2016learning} and two kinds of approaches gain attraction.
%and beyond these simple variants of LSTMs, two kinds of approaches are adopted to cope with the deficiency of internal memory.
One is to augment LSTM with an external memory \cite{sukhbaatar2015end,tran2016recurrent}, but they are of poor performance on time because of the huge external memory matrix.
% is supposed to be scanned every time-step.
Unlike these methods, we fully exploit the potential of internal memory of LSTM by adjusting its forgetting rates.

The other one tries to use multiple time-scales to distinguish different states \cite{el1995hierarchical,Koutnik:2014vy,liu2015multitimescale}. They partition the hidden states into several groups and each group is activated and updated at different frequencies (e.g. one group updates every 2 time-step, the other updates every 4 time-step). In these methods, different memory groups are not fully interconnected, and the information is transmitted from faster groups to slower ones, or vice versa.

However, the memory of slower groups are not updated at every step, which may lead to sentiment information loss and semantic inconsistency. In our proposed CLSTM, we assign different forgetting rates to memory groups. This novel strategy enable each memory group to be updated at every time-step, and every bit of the long-term and short-term memories in previous time-step to be taken into account when updating.

%To address this problem, \newcite{liu2015multi} apply clockwork recurrent neural networks (CW-RNN) \cite{koutnik2014clockwork} to better model long sentences, and achieve great success. Concretely, CW-RNN separates the LSTM units into several groups. Different
%groups capture different timescales dependencies and updated with the different frequencies. The fast-speed (updated frequently) groups are short-term memories, while the slow-speed (updated infrequently) groups are long-term memories. However, CW-RNN could only accumulate some valuable information from the short-term memory into the long-term memory or vice versa, which is so different with the human memory. %We would update long or short term memory by incorporating all memory
%Besides, not all the memory groups of CW-RNN would update at each time step, which are updated according to their updating frequencies respectively, which is not reasonable.

%To better model the multiple time scale memories, in this work, we propose MTS-LSTM by constraining the scales of forget gate values in different groups. All groups of MTS-LSTM are updated at each time step, and the update of each group is based on the all groups at previous time step with different time scale memories.

\section{Long Short-Term Memory Networks}

%Recurrent Neural Networks (RNNs) suffer from the notorious gradient vanishing and exploding problems due to the long-range back-propagation gradient though time, and LSTM is a successful adaptation to prolong the memory capacity by adding memory cells and activation gating, but further efforts have been made to eliminate the gradient problem.

Long short-term memory network (LSTM) \cite{hochreiter1997long} is a typical recurrent neural network, which alleviates the problem of gradient diffusion and explosion. LSTM can capture the long dependencies in a sequence by introducing a memory unit and a gate mechanism which aims to decide how to utilize and update the information kept in memory cell.

%The memory cell $\bc$ is controlled by three gates: input gate $\mathbf{i}$, forget gate $\mathbf{f}$ and output
%gate $\mathbf{o}$:
Formally, the update of each LSTM component can be formalized as:

\begin{align}
    \mathbf{i}^{(t)} &= \sigma (\bW_{i} \x^{(t)} + \bU_{i} \bh^{(t-1)}), \\
    \mathbf{f}^{(t)} &= \sigma (\bW_{f} \x^{(t)} + \bU_{f} \bh^{(t-1)}), \\
    \mathbf{o}^{(t)} &= \sigma (\bW_{o} \x^{(t)} + \bU_{o} \bh^{(t-1)}), \\
    \tilde{\bc}^{(t)} &=  \tanh (\bW_{c} \x^{(t)} + \bU_{c} \bh^{(t-1)}), \\ \label{eq:memory_cell}
    \bc^{(t)} &= \mathbf{f}^{(t)} \odot \bc^{(t-1)} + \mathbf{i}^{(t)} \odot \tilde{\bc}^{(t)}, \\
    %\bc^{(t)} &= \mathbf{f}^{(t)} \odot \bc^{(t-1)} + \mathbf{i}^{(t)} \odot \phi(\bW_{cx} \x^{(t)} + \bW_{ch} \bh^{(t-1)}), \\
    \bh^{(t)} &= \mathbf{o}^{(t)} \odot \tanh (\bc^{(t)}),
\end{align}
where $\sigma$ is the logistic sigmoid function. Operator $\odot$ is the element-wise multiplication operation.
$\mathbf{i}^{(t)}$, $\mathbf{f}^{(t)}$, $\mathbf{o}^{(t)}$ and $\mathbf{c}^{(t)}$ are
the input gate, forget gate, output gate, and memory
cell activation vector at time-step $t$ respectively, all of which have the same size
as the hidden vector $\h^{(t)} \in \mathbb{R}^{H}$.
$\bW_{i}$, $\bW_{f}$, $\bW_{o} \in \mathbb{R}^{H \times d}$ and $\bU_{i}$, $\bU_{f}$, $\bU_{o} \in \mathbb{R}^{H \times H}$ are trainable parameters. Here, $H$ and $d$ are the dimensionality of hidden layer and input respectively.

\begin{figure}[t]\centering
  \includegraphics[width=0.85\linewidth]{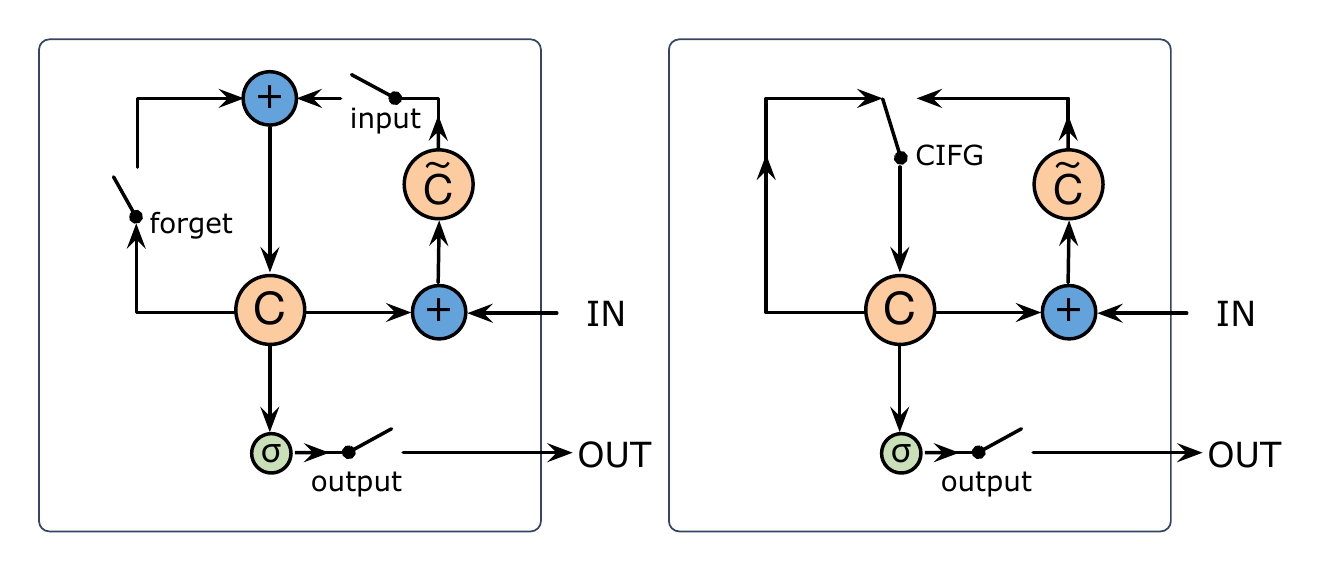}
  \caption{(a) A standard LSTM unit and (b) a CIFG-LSTM unit. There are three gates in (a), the input gate, forget gate and output gates, while in (b), there are only two gates, the CIFG gate and output gate. }\label{fig:CIFG}
\end{figure}
%\footnote{the firgue is depicted in the fashion of \cite{chung2014empirical}}
%
%
%The LSTM transition equations are the following:

\section{Cached Long Short-Term Memory Neural Network}

%\danlu{this part should be edited}

LSTM is supposed to capture the long-term and short-term dependencies simultaneously, but when dealing with considerably long texts, LSTM  also fails on capturing and understanding significant sentiment message \cite{le2015simple}. Specifically, the error signal would nevertheless suffer from gradient vanishing in modeling long texts with hundreds of words and thus the network is difficult to train.
% using back-propagation through time (BPTT) \cite{werbos1990backpropagation} algorithm.

%inevitably loss information in the long transmission process \cite{

Since the standard LSTM inevitably loses valuable features, we propose a cached long short-term memory neural networks (CLSTM) to capture information in a longer steps by introducing a cache mechanism.
%CLSTM introduces a cache mechanism, which divides memory into several groups and assigns different forgetting rates, as filters, to different groups.
Moreover, in order to better control and balance the historical message and the incoming information, we adopt one particular variant of LSTM proposed by \newcite{Greff:2015wv}, the Coupled Input and Forget Gate LSTM (CIFG-LSTM).

\paragraph{Coupled Input and Forget Gate LSTM}

Previous studies show that the merged version gives performance comparable to a standard LSTM on language modeling and classification tasks because using the input gate and forget gate simultaneously incurs redundant information \cite{chung2014empirical,Greff:2015wv}.

In the CIFG-LSTM, the input gate and forget gate are coupled as one uniform gate, that is, let $\mathbf{i}^{(t)}$ = $\mathbf{1} - \mathbf{f}^{(t)}$. We use $\mathbf{f}^{(t)}$ to denote the coupled gate. Formally, we will replace Eq. \ref{eq:memory_cell} as below:
\begin{equation}
  \bc^{(t)} = \mathbf{f}^{(t)} \odot \bc^{(t-1)} + (\mathbf{1} - \mathbf{f}^{(t)}) \odot \tilde{\bc}^{(t)} \\
\end{equation}

Figure \ref{fig:CIFG} gives an illustrative comparison of a standard LSTM and the CIFG-LSTM.

%To the best of our knowledge, our approach is the first to group the neurons and force them to learn separately different range of memory.

\paragraph{Cached LSTM}

Cached long short-term memory neural networks (CLSTM) aims at capturing the long-range information by a cache mechanism, which divides memory into several groups, and different forgetting rates, regarded as filters, are assigned to different groups.
%Specifically, the faster groups with faster forgetting rate play the role of a cache, transiting information from faster groups to slower groups.

Different groups capture different-scale dependencies by squashing the scales of forgetting rates. The groups with high forgetting rates are short-term memories, while the groups with low forgetting rates are long-term memories.

Specially, we divide the memory cells into $K$ groups $\{G_1,\cdots,G_K\}$. Each group includes a internal memory $\mathbf{c}_k$, output gate $\mathbf{o}_k$ and forgetting rate $\mathbf{r}_k$. The forgetting rate of different groups are squashed in distinct ranges.

We modify the update of a LSTM as follows.

\begin{align}
 \mathbf{r}_k^{(t)} &= \psi_k\left(\sigma(\bW^k_r\x^{(t)}+\sum_{j=1}^K\bU^{j\rightarrow k}_f\h_j^{(t-1)})\right), \label{eq:mtlstm-eqs-2}  \\
 \mathbf{o}_k^{(t)} &=\sigma(\bW^k_o\x^{(t)}+\sum_{j=1}^K\bU^{j\rightarrow k}_o\h_j^{(t-1)}),  \label{eq:mtlstm-eqs-3} \\
 \tilde{\mathbf{c}}{_k^{(t)}} &=\tanh(\bW^k_c\x^{(t)}+\sum_{j=1}^K\bU^{j\rightarrow k}_c\h_j^{(t-1)}), \label{eq:mtlstm-eqs-4} \\
\mathbf{c}_k^{(t)} &=(1-\mathbf{r}_k^{(t)}) \odot \mathbf{c}_k^{(t-1)} + (\mathbf{r}_k^{(t)}) \odot \tilde{\mathbf{c}}{_k^{(t)}}, \label{eq:mtlstm-eqs-5} \\
\h_k^{(t)} &=\mathbf{o}_k^{(t)} \odot \tanh(\mathbf{c}_k^{(t)}) \label{eq:mtlstm-eqs-6},
\end{align}
where $\mathbf{r}_k^{(t)}$ represents forgetting rate of the $k$-th memory group at step $t$; $\psi_k$ is a squash function, which constrains the value of forgetting rate $\mathbf{r}_k$ within a range. To better distinguish the different role of each group, its forgetting rate is squashed into a distinct area. The squash function $\psi_k(\bz)$ could be formalized as:
\begin{align}
\mathbf{r}_k = \psi_k(\bz) =\frac{1}{K} \cdot \bz + \frac{k-1}{K},
\end{align}
where $\bz \in (0,1)$ is computed by logistic sigmoid function. Therefore, $\mathbf{r}_k$ can constrain the forgetting rate in the range of $(\frac{k-1}{K},\frac{k}{K})$.

Intuitively, if a forgetting rate $\mathbf{r}_k$  approaches to $\mathbf{0}$, the group $k$ tends to be the long-term memory; if a $\mathbf{r}_k$  approaches to $\mathbf{1}$, the group $k$ tends to be the short-term memory.  Therefore, group $G_1$ is the slowest, while group $G_K$ is the fastest one. The faster groups are supposed to play a role as a cache, transiting information from faster groups to slower groups.

% for division of labour of memories in different timescales. Thus, we investigate two time scale functions.
%Therefore, the value scale of $\mathbf{r}$ of each groups is crucial.

%and controls the different timescales of memory groups.
\begin{figure}[t]\centering
  \includegraphics[width=0.85\linewidth]{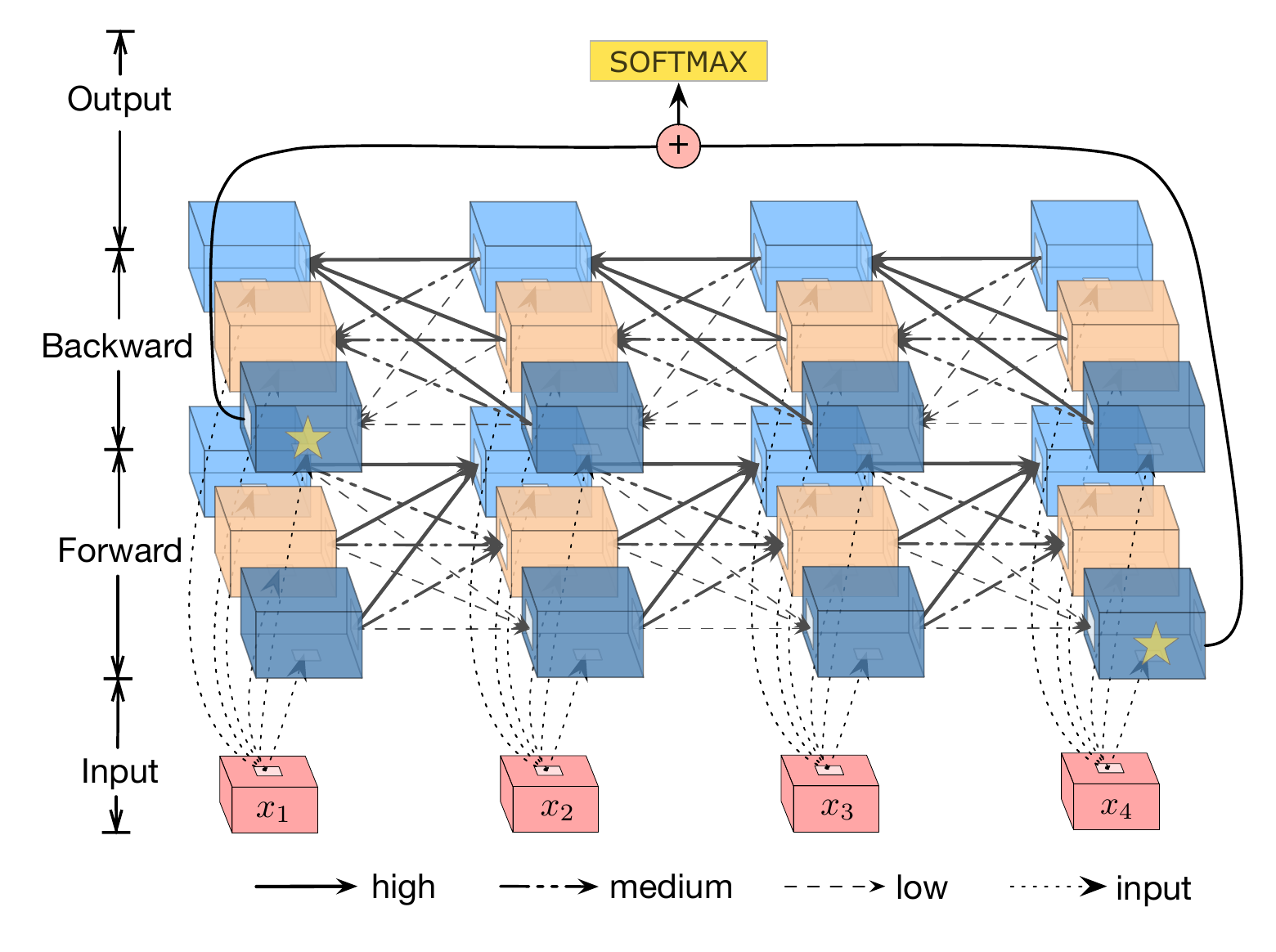}
  \caption{An overview of the proposed architecture. Different styles of arrows indicate different forgetting rates. Groups with stars are fed to a fully connected layers for softmax classification. Here is an instance of B-CLSTM with text length equal to $4$ and the number of memory groups is $3$. }\label{fig:update}
\end{figure}

\paragraph{Bidirectional CLSTM}
\newcite{graves2005framewise} proposed a Bidirectional LSTM (B-LSTM) model, which utilizes additional backward information and thus enhances the memory capability.

We also employ the bi-directional mechanism on CLSTM and words in a text will receive information from both sides of the context. Formally, the outputs of forward LSTM for the $k$-th group is $[\overrightarrow{\mathbf{h}}_k^{(1)},\overrightarrow{\mathbf{h}}_k^{(2)},\dots,\overrightarrow{\mathbf{h}}_k^{(T)}]$. The outputs of backward LSTM for the $k$-th group is $[\overleftarrow{\mathbf{h}}_k^{(1)},\overleftarrow{\mathbf{h}}_k^{(2)},\dots,\overleftarrow{\mathbf{h}}_k^{(T)}]$.

Hence, we encode each word $w_t$ in a given text $w_{1:T}$ as $\mathbf{h}_k^{(t)}$:
\begin{equation}
  \mathbf{h}_k^{(t)} = {\overrightarrow{\mathbf{h}}_k^{(t)}} \oplus {\overleftarrow{\mathbf{h}}_k^{(t)}},\\
\end{equation}
where the $\oplus$ indicates concatenation operation.

\begin{table*}[t]\small\centering
  \begin{tabular}{|*{8}{c|}}
    \hline
    % after \\: \hline or \cline{col1-col2} \cline{col3-col4} ...
    Dataset & Type & Train Size & Dev. Size& Test Size & Class & Words/Doc & Sents/Doc  \\
    \hline
    IMDB & Document     & 67426 & 8381 & 9112   & 10 & 394.6 & 16.08\\
Yelp 2013& Document     & 62522 & 7773 & 8671   & 5 & 189.3 & 10.89\\
    Yelp 2014 & Document    & 183019 & 22745 & 25399& 5  & 196.9 & 11.41\\
    \hline
  \end{tabular}
  \caption{Statistics of the three datasets used in this paper. The rating scale (Class) of Yelp2013 and Yelp2014 range from 1 to 5 and that of IMDB ranges from 1 to 10. Words/Doc is the average length of a sample and Sents/Doc is the average number of sentences in a document. }\label{tab:data}
\end{table*}

%\paragraph{Time Scale Function - I}
%Time scale function $\psi_k^{\text{I}} (\cdot)$ tries to assign groups in hierarchical architecture, where show group also tries to cover the short term dependency:
%\begin{align}
%\mathbf{f}_k &= \psi_k^{\text{I}}(\tilde{\mathbf{f}}_k)\notag\\
%&=k \cdot \xi \cdot \tilde{\mathbf{f}}_k,\\
%\xi &= 1/K,
%\end{align}
%
%Concretely, if $k=4$, we would constrains the value of $\mathbf{f}$ in different groups as:
%\begin{align}
%  \mathbf{f}_1 &\in [0,0.25)\\
%    \mathbf{f}_2 &\in [0.25,0.50)\\
%      \mathbf{f}_3 &\in [0.50,0.75)\\
%        \mathbf{f}_4 &\in [0.75,1.00]\\
%\end{align}

%\paragraph{Time Scale Function - II}
%Time scale function $\psi_k^{\text{II}} (\cdot)$ tries to assign groups in parallel architecture:%, where show group also tries to cover the short term dependency:
%\begin{align}
%\mathbf{f}_k &= \psi_k^{\text{II}}(\tilde{\mathbf{f}}_k)\notag\\
%&=\xi \cdot \tilde{\mathbf{f}}_k + (k-1) \cdot \xi,\\
%\xi &= 1/K,
%\end{align}

\paragraph{Task-specific Output Layer for Document-level Sentiment Classification}

With the capability of modeling long text, we can use our proposed model to analyze sentiment in a document. Figure \ref{fig:update} gives an overview of the architecture.

%The first step of is to map words into distributed vectors, namely word embeddings.
%The embedding layer can be regarded as a simple projection layer. For a given sentence $w_{1:T}$, we would receive the word embeddings $[x^{(1)}, x^{(2)}, \dots,x^{(T)}]$ by looking up embedding matrix $\bM$ according to their indices, where $T$ is the length of the sentence.
%With the word embeddings in a document, we can get it distributed representation by CLSTM or B-CLSTM.

Since the first group, the slowest group, is supposed to keep the long-term information and can better represent a whole document, we only utilize the final state of this group to represent a document. As for the B-CLSTM, we concatenate the state of the first group in the forward LSTM at $T$-th time-step and the first group in the backward LSTM at first time-step.

Then, a fully connected layer followed by a softmax function is used to predict the probability distribution over classes for a given input. Formally, the probability distribution $\mathbf{p}$ is:
\begin{equation}
  \mathbf{p} = \softmax (\bW_p\times \bz + \bb_p),\\
\end{equation}
where $\bW_p$ and $\bb_p$ are model's parameters. Here $\bz$ is $\overrightarrow{\mathbf{h}}^{(T)}_1$ in CLSTM, and $\bz$ is ${[{\overrightarrow{\mathbf{h}}^{(T)}_1} \oplus {\overleftarrow{\mathbf{h}}^{(1)}_1}]}$ in B-CLSTM.

%\paragraph{Embedding Layer}

%Formally, we have a word dictionary $\mathcal{D}$ of size $|\mathcal{D}|$, which is extracted from the training set and unknown words (out of vocabulary words) are mapped to a special symbol that is not used elsewhere. Hence, each word $w \in \mathcal{W}$ is represented as a vector (word embedding) $\bv_w \in \mathbb{R}^d$, where $d$ is the dimensionality of the vector space. Then the character embeddings are stacked into an embedding matrix $\bM \in \mathbb{R}^{d \times |\mathcal{D}|}$.
%
%%For a word $w \in \mathcal{D}$, the corresponding word embedding $\bv_w \in \mathbb{R}^d$ is retrieved from the embedding matrix $\bM$. %Intuitively, the embedding layer can be regarded as a simple projection layer
%%where the word embedding $\bv_w$ for each word $w_i$ in a given sentence $w_{1:n}$ is achieved by table lookup operation according to its index, where $n$ is the length of the sentence.
%Intuitively, the embedding layer can be regarded as a simple projection layer. For a given sentence $w_{1:T}$, we would receive the word embeddings $[x^{(1)}, x^{(2)}, \dots,x^{(T)}]$ by looking up embedding matrix $\bM$ according to their indices, where $T$ is the length of the sentence.

%\paragraph{CLSTM Layer}

\section{Training}

The objective of our model is to minimize the cross-entropy error of the predicted and true distributions. Besides, the objective includes an $L_2$ regularization
term over all parameters. Formally, suppose we have $m$ train sentence and label pairs $(w^{(i)}_{1:T_i},y^{(i)})_{i=1}^m$, the object is to minimize the objective function $J(\theta)$:
\begin{equation}
  J(\theta) =- \frac{1}{m} \sum_{i=1}^m \log \mathbf{p}^{(i)}_{y^{(i)}} + \frac{\lambda}{2} || \theta||^2,
\end{equation}
where $\theta$ denote all the trainable parameters of our model.

%\paragraph{Optimizer}
%To optimize the objective function, we adopt Adagrad update rule \cite{duchi2011adaptive} with mini-batches.
%The parameter update for the $i$-th
%parameter $\theta_{t,i}$ at time step $t$ is as follows:
%\begin{equation}
% \theta_{t, i} = \theta_{t-1, i} - \frac{\alpha}{\sqrt{\sum\nolimits_{\tau=1}^{t}g_{\tau,i}^2}}g_{t, i},
%\end{equation}
%where $\alpha$ is the initial learning rate and $g_\tau \in \mathbb{R}^{|\theta_{\tau,i}|}$ is the gradient at time step $\tau$ for parameter $\theta_{\tau,i}$.

\section{Experiment}
In this section, we study the empirical result of our model on three datasets for document-level sentiment classification. Results show that the proposed model outperforms competitor models from several aspects when modelling long texts.

\subsection{Datasets}
Most existing datasets for sentiment classification such as Stanford Sentiment Treebank \cite{socher2013recursive} are composed of short paragraphs with several sentences, which cannot evaluate the effectiveness of the model under the circumstance of encoding long texts. We evaluate our model on three popular real-world datasets, Yelp 2013, Yelp 2014 and IMDB. Table \ref{tab:data} shows the statistical information of the three datasets. All these datasets can be publicly accessed\footnote{\url{http://ir.hit.edu.cn/~dytang/paper/acl2015/dataset.7z}}. We pre-process and split the datasets in the same way as \newcite{Tang:2015vb} did.

\begin{itemize*}
  \item \textbf{Yelp 2013} and \textbf{Yelp 2014} are review datasets derived from Yelp Dataset Challenge\footnote{\url{http://www.yelp.com/dataset_challenge}} of year 2013 and 2014 respectively. The sentiment polarity of each review is 1 star to 5 stars, which reveals the consumers' attitude and opinion towards the restaurants.
%   \item \textbf{Yelp 2014} is a customer review dataset derived from Yelp Dataset Challenge of year 2014. It scales from 1 to 5.
  \item \textbf{IMDB}  is a popular movie review dataset consists of 84919 movie reviews ranging from 1 to 10. Average length of each review is 394.6 words, which is much larger than the length of two Yelp review datasets.
%   \cite{DBLP:conf/kdd/DiaoQWSJW14}
\end{itemize*}

\subsection{Evaluation Metrics}
We use Accuracy (Acc.) and MSE as evaluation metrics for sentiment classification. Accuracy is a standard metric to measure the overall classification result and Mean Squared Error (MSE) is used to figure out the divergences between predicted sentiment labels and the ground truth ones. %  Accuracy only reveals the hit rate but MSE could reflect the distance between prediction and ground truth.
% \begin{equation}
% \text{MSE} = \frac{\sum\nolimits_{i}^{N}\text{gold}_{i}-\text{predicted}_{i}}{N},
% \end{equation}
% where $N$ is the number of samples.
% \begin{table*}[t]
\begin{table*}[t]\small
\center
\begin{tabular}{|l|*{6}{c|}}
\hline
\multirow{2}*{\textbf{Model}} & \multicolumn{2}{c|}{IMDB}    &   \multicolumn{2}{c|}{Yelp 2014} &\multicolumn{2}{c|}{Yelp 2013} \\\cline{2-7}
&Acc. (\%)&MSE&Acc. (\%)&MSE&Acc. (\%)&MSE\\\hline
% Majority & 19.6 & 6.225 & 39.2 & 1.203 & 41.1 & 1.12 \\
CBOW    &   34.8&  2.867 & 56.8 &  0.620  &  54.5&  0.706\\
PV \cite{Tang:2015vb}& 34.1&3.291 & 56.4 & 0.643 &55.4 & 0.692 \\
RNTN+Recurrent \cite{Tang:2015vb} &40.0&3.112&58.2&0.674&57.4&0.646\\
UPNN (CNN) \cite{Tang:2015vb}&40.5&2.654&58.5&0.653&57.7&0.659\\
JMARS* \cite{Diao:2014gn} &-&3.143&-&0.998&-&0.970\\
UPNN (CNN)* \cite{Tang:2015vb} &   \textbf{43.5}    &  \textbf{2.566}& \textbf{60.8}   &   \textbf{0.584}  &   \textbf{59.6}   &\textbf{0.615}\\
\hline
RNN     &20.5   &6.163  &   41.0&  1.203    &42.8 & 1.144 \\
LSTM    & 37.8  & 2.597  &   56.3   &  0.592 & 53.9&  0.656  \\
CIFG-LSTM   &   39.1   &   2.467  &   55.2   &  0.598 & 57.3    & 0.558  \\
CLSTM   &   \textbf{42.1}   &  \textbf{2.399} &  \textbf{59.2} &  \textbf{0.539} & \textbf{59.4}  &   \textbf{0.587}   \\
\hline
BLSTM  &43.3   & 2.231  &   59.2    & 0.538 & 58.4  &   0.583   \\
CIFG-BLSTM & 44.5 &    2.283 &   60.1    & 0.527 & 59.2  &   0.554   \\
B-CLSTM   & \textbf{46.2} & \textbf{2.112} & \textbf{61.9} & \textbf{0.496} & \textbf{59.8} &   \textbf{0.549}\\

% Multi-BiLSTM\\
\hline
\end{tabular}
\caption{Sentiment classification results of our model against competitor models on IMDB, Yelp 2014 and Yelp 2013. Evaluation metrics are classification accuracy (Acc.) and MSE. Models with * use user and product information as additional features. Best results in each group are in bold. }\label{tab:results}
\end{table*}

 \begin{table}[t] \setlength{\tabcolsep}{3pt}
 \centering
 \begin{tabular}{|l|*{3}{c|}}
     \hline
     Dataset& IMDB & Yelp13 & Yelp14\\\hline
     Hidden layer units &120 & 120 &120\\
     Number of groups &3&4&4\\
     Weight Decay & $1\mathrm{e}{-4}$ & $1\mathrm{e}{-4}$ & $5\mathrm{e}{-4}$\\
     Batch size & 128&64&64\\
     \hline
 \end{tabular}
 \caption{Optimal hyper-parameter configuration for three datasets.}\label{tab:paramSet}
 \end{table}

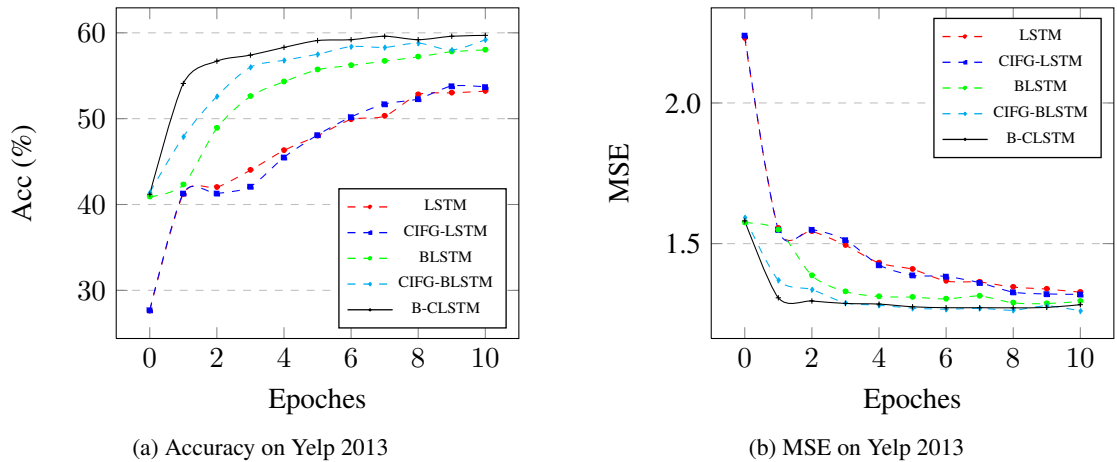
\begin{figure*}[t]
  \centering
  \pgfplotsset{width=0.42\textwidth}
  \subfloat[Accuracy on Yelp 2013]{
  \begin{tikzpicture}
    \begin{axis}[
%    title=Inv. cum. normal,
    xlabel={Epoches},
    ylabel={Acc (\%)},
    legend entries={LSTM,CIFG-LSTM, BLSTM,  CIFG-BLSTM,B-CLSTM},
    mark size=1.0pt,
    %no markers,
    ymajorgrids=true,
    grid style=dashed,
    %cycle list name=exotic,
   %cycle list name=my black white,
    legend pos= south east,%
    legend style={font=\tiny,line width=.5pt,mark size=.6pt,%
            legend columns=1,
            /tikz/every even column/.append style={column sep=0.5em}},
            smooth,
    ]
    \addplot [red,dashed,mark=*] table [x index=0, y index=1] {conv_speed_LSTM.txt};
    \addplot [blue,dashed,mark=square*] table [x index=0, y index=1] {conv_speed_LSTM_s.txt};
   \addplot [green,dashed,mark=otimes*] table [x index=0, y index=1] {conv_speed_BLSTM.txt};
    \addplot [cyan,dashed,mark=diamond*] table [x index=0, y index=1] {conv_speed_BLSTM_s.txt};
    \addplot [black,solid,mark=+] table [x index=0, y index=1] {conv_speed_Multi_BLSTM.txt};
    %%\addplot [blue,dashed,mark=square*] table [x index=0, y index=2] {axis/model_select.txt};
%    \addplot [green,dashed,mark=otimes*] table [x index=0, y index=3] {axis/model_select.txt};
%    %\addplot [cyan,dashed,mark=diamond*] table [x index=0, y index=4] {axis/model_select.txt};
%   \addplot [pink,densely dashed,mark=triangle*] table [x index=0, y index=5] {axis/model_select.txt};
%    %\addplot [black,solid,mark=+] table [x index=0, y index=6] {axis/model_select.txt};
%    \addplot [yellow,solid,mark=*] table [x index=0, y index=7] {axis/model_select.txt};
%    %\addplot [purple,solid,mark=square*] table [x index=0, y index=8] {axis/model_select.txt};
%    %\addplot [red,dotted,mark=*] table [x index=0, y index=1] {axis/LayerWise_2_dev.txt};
%    %\addplot [blue,dotted,mark=*] table [x index=0, y index=1] {axis/LayerWise_3_dev.txt};
%    %\addplot [green,dotted,mark=*] table [x index=0, y index=1] {axis/LayerWise_4_dev.txt};
%    %\addplot [pink,dotted,mark=*] table [x index=0, y index=1] {axis/LayerWise_5_dev.txt};
%    %\addplot table [x=k, y=baseline] {rerank-dev.txt};
    \end{axis}
\end{tikzpicture}
}
  \hspace{2em}
\subfloat[MSE on Yelp 2013]{
  \begin{tikzpicture}
    \begin{axis}[
%    title=Inv. cum. normal,
    xlabel={Epoches},
    yticklabels={0.5, 1.0, 1.5, 2.0},
    ylabel={MSE},
    legend entries={LSTM,CIFG-LSTM, BLSTM,  CIFG-BLSTM,B-CLSTM},
    mark size=1.0pt,
    %no markers,
    ymajorgrids=true,
    grid style=dashed,
    %cycle list name=exotic,
   %cycle list name=my black white,
    legend pos= north east,%
    legend style={font=\tiny,line width=.5pt,mark size=.6pt,%
            legend columns=1,
            /tikz/every even column/.append style={column sep=0.5em}},
            smooth,
    ]
    \addplot [red,dashed,mark=*] table [x index=0, y index=2] {conv_speed_LSTM.txt};
    \addplot [blue,dashed,mark=square*] table [x index=0, y index=2] {conv_speed_LSTM_s.txt};
    \addplot [green,dashed,mark=otimes*] table [x index=0, y index=2] {conv_speed_BLSTM.txt};
    \addplot [cyan,dashed,mark=diamond*] table [x index=0, y index=2] {conv_speed_BLSTM_s.txt};
    \addplot [black,solid,mark=+] table [x index=0, y index=2] {conv_speed_Multi_BLSTM.txt};
    \end{axis}
\end{tikzpicture}
}
  \caption{Convergence speed experiment on Yelp 2013. X-axis is the iteration epoches and Y-axis is the classifcication accuracy(\%) achieved.}
  \label{fig:conv}
\end{figure*}

\subsection{Baseline Models}
We compare our model, CLSTM and B-CLSTM with the following baseline methods.

\begin{itemize*}
  \item \textbf{CBOW} sums the word vectors and applies a non-linearity followed by a softmax classification layer.
    % \item \textbf{RNTN+Recurrent} represents sentence with RNTN \cite{Socher:2013tz} and compose document level representation with recurrent neural network.
     \item \textbf{JMARS} is one of the state-of-the-art recommendation algorithm, which leverages user and aspects of a review with collaborative filtering and topic modeling.
  \item \textbf{CNN} UPNN (CNN) \cite{Tang:2015vb} can be regarded as a CNN \cite{kim2014convolutional}. Multiple filters are sensitive to capture different semantic features during generating a representation in a bottom-up fashion.
   \item \textbf{RNN} is a basic sequential model to model texts \cite{Elman:1991fu}.
  \item \textbf{LSTM} is a recurrent neural network with memory cells and gating mechanism \cite{hochreiter1997long}.
  \item \textbf{BLSTM} is the bidirectional version of LSTM, and can capture more structural information and longer distance during looking forward and back \cite{graves2013hybrid}.
  \item \textbf{CIFG-LSTM \& CIFG-BLSTM} are Coupled Input Forget Gate LSTM and BLSTM, denoted as CIFG-LSTM and CIFG-BLSTM respectively \cite{Greff:2015wv}. They combine the input and forget gate of LSTM and require smaller number of parameters in comparison with the standard LSTM.
\end{itemize*}

\subsection{Hyper-parameters and Initialization}
For parameter configuration, we choose parameters on validation set mainly according to classification accuracy for convenience because MSE always has strong correlation with accuracy. The dimension of pre-trained word vectors is 50.  We use 120 as the dimension of hidden units, and choose weight decay among \{ $5\mathrm{e}{-4}$, $1\mathrm{e}{-4}$, $1\mathrm{e}{-5}$ \}. We use Adagrad \cite{Duchi:2011wu} as optimizer and its initial learning rate is 0.01. Batch size is chosen among \{ 32, 64, 128 \} for efficiency. For CLSTM, the number of memory groups is chosen upon each dataset, which will be discussed later.
We remain the total number of the hidden units unchanged. Given 120 neurons in all for instance, there are four memory groups and each of them has 30 neurons. This makes model comparable to (B)LSTM. Table \ref{tab:paramSet} shows the optimal hyper-parameter configurations for each dataset.

%with orthogonal initialization \cite{Saxe:2013tq,DBLP:journals/corr/HenaffSL16} while non-recurrent parameters are initialized
For model initialization, we initialize all recurrent matrices with randomly sampling from uniform distribution in [-0.1, 0.1].
Besides, we use GloVe\cite{Pennington:2014uw} as pre-trained word vectors. The word embeddings are fine-tuned during training. Hyper-parameters achieving best results on the validation set are chosen for final evaluation on test set.

% \subsection{Model Selection}

\subsection{Results}
The classification accuracy and mean square error (MSE) of our models compared with other competitive models are shown in Table \ref{tab:results}. When comparing our models to other neural network models, we have several meaningful findings.

\begin{enumerate*}

  \item Among all unidirectional sequential models, RNN fails to capture and store semantic features while vanilla LSTM preserves sentimental messages much longer than RNN. It shows that internal memory plays a key role in text modeling. CIFG-LSTM gives performance comparable to vanilla LSTM.

  \item With the help of bidirectional architecture, models could look backward and forward to capture features in long-range from global perspective. In sentiment analysis, if users show their opinion at the beginning of their review, single directional models will possibly forget these hints.

%  \item We replace the (B)LSTM with a CIFG-(B)LSTM mainly for two reasons:
%  \begin{enumerate*}
%     \item (B)LSTM with scale function may lead to unstableness in numeric value when scaling input and forget gates.
%     \item According to \cite{Chung:2014wf,Greff:2015wv}, the input gate and forget gate in LSTM contain redundant information so we could simplify them. CIFG-(B)LSTM has smaller number of parameter than counterpart (B)LSTM. \TODO{move this part to model/related work}
%
%  \end{enumerate*}
  \item The proposed CLSTM beats the CIFG-LSTM and vanilla LSTM and even surpasses the bidirectional models. In Yelp 2013, CLSTM achieves 59.4\% in accuracy, which is only 0.4 percent worse than B-CLSTM, which reveals that the cache mechanism has successfully and effectively stored valuable information without the support from bidirectional structure.

  \item Compared with existing best methods, our model has achieved new state-of-the-art results by a large margin on all document-level datasets in terms of classification accuracy. Moreover, B-CLSTM even has surpassed JMARS and CNN (UPNN) methods which utilized extra user and product information.

  \item In terms of time complexity and numbers of parameters, our model keeps almost the same as its counterpart models while models of hierarchically composition may require more computational resources and time.

\end{enumerate*}

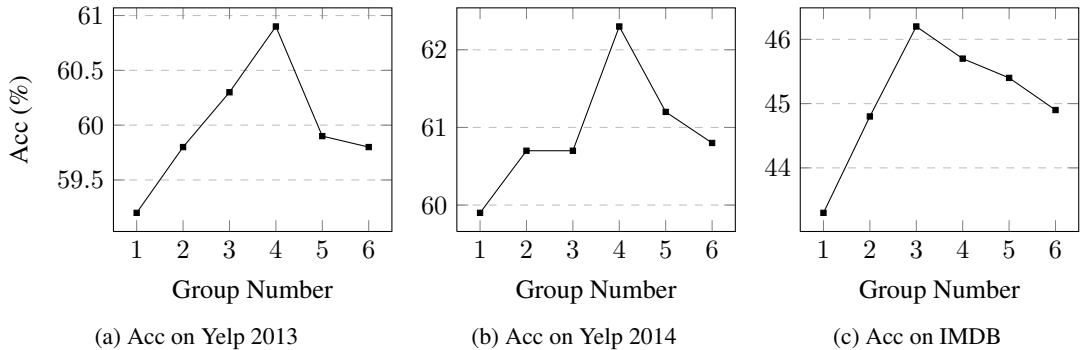
\begin{figure*}[t]\small
 \centering
\pgfplotsset{width=0.32\textwidth}
\subfloat[Acc on Yelp 2013]{
 \begin{tikzpicture}
  \begin{axis}[
  xlabel={Group Number},
  ylabel={Acc (\%)},
  xtick={1,2,3,4,5,6},
  mark size=1.0pt,
  ymajorgrids=true,
  grid style=dashed,
%      smooth,
  ]
 \addplot [black,solid,mark=square*] table [x index=0, y index=1] {MEM2013.txt};

  \end{axis}
\end{tikzpicture}
}
\hspace{0em}
\subfloat[Acc on Yelp 2014]{
 \begin{tikzpicture}
  \begin{axis}[
 xlabel={Group Number},
  xtick={1,2,3,4,5,6},
  mark size=1.0pt,
  ymajorgrids=true,
  grid style=dashed,
%      smooth,
  ]
 \addplot [black,solid,mark=square*] table [x index=0, y index=1] {MEM2014.txt};

  \end{axis}
\end{tikzpicture}
}
\hspace{0em}
\subfloat[Acc on IMDB]{
 \begin{tikzpicture}
  \begin{axis}[
  xlabel={Group Number},
  xtick={1,2,3,4,5,6},
  mark size=1.0pt,
  ymajorgrids=true,
  grid style=dashed,
%      smooth,
  ]
 \addplot [black,solid,mark=square*] table [x index=0, y index=1] {MEMIMDB.txt};

  \end{axis}
\end{tikzpicture}
}
%
%\hspace{0em}
%\subfloat[MSE of Yelp2013.]{
% \begin{tikzpicture}
%  \begin{axis}[
%  xlabel={epoches},
%  ylabel={MSE},
%  mark size=1.0pt,
%  ymajorgrids=true,
%  grid style=dashed,
%%      smooth,
%  ]
% \addplot [blue,solid,mark=square*] table [x index=0, y index=2] {MEM2013.txt};
%
%  \end{axis}
%\end{tikzpicture}
%}
%\hspace{0em}
%\subfloat[MSE of Yelp2014.]{
% \begin{tikzpicture}
%  \begin{axis}[
%  xlabel={epoches},
%  ylabel={MSE},
%  mark size=1.0pt,
%  ymajorgrids=true,
%  grid style=dashed,
%%      smooth,
%  ]
% \addplot [blue,solid,mark=square*] table [x index=0, y index=2] {MEM2014.txt};
%
%  \end{axis}
%\end{tikzpicture}
%}
%\hspace{0em}
%\subfloat[MSE of IMDB.]{
% \begin{tikzpicture}
%  \begin{axis}[
%  xlabel={epoches},
%  ylabel={MSE},
%  mark size=1.0pt,
%  ymajorgrids=true,
%  grid style=dashed,
%%      smooth,
%  ]
% \addplot [blue,solid,mark=square*] table [x index=0, y index=2] {MEMIMDB.txt};
%
%  \end{axis}
%\end{tikzpicture}
%}
  \caption{Classification accuracy on different number of memory group on three datasets. X-axis is the number of memory group(s). }
   \label{fig:mem}
\end{figure*}

\begin{figure}[t]\small
  \centering
  \pgfplotsset{width=0.47\textwidth}
  \begin{tikzpicture}
    \begin{axis}[
%    title=Inv. cum. normal,
    xlabel={Length Ranking (\%)},
    xtick=data,
    xticklabels={10,20,30,40,50,60,70,80,90,100},
    ymax=50,
    ylabel={Acc (\%)},
    legend entries={CBOW, CIFG-LSTM,CLSTM,  CIFG-BLSTM,  B-CLSTM},
    mark size=1.0pt,
    %no markers,
    ymajorgrids=true,
    grid style=dashed,
    y label style={at={(.05,0.55)}},
    %cycle list name=exotic,
   %cycle list name=my black white,
    % legend pos= north east,%
    legend style={font=\tiny,line width=.3pt,mark size=.5pt,%
            at={(1.0,-0.4)},
            legend columns=3,
            anchor=south east,
            /tikz/every even column/.append style={column sep=0.5em}},
%            smooth,
    ]
    \addplot [green,solid,mark=*] table [x index=0, y index=1] {length.txt};
    \addplot [blue,solid,mark=square*] table [x index=0, y index=2] {length.txt};
    \addplot [purple,solid,mark=otimes*] table [x index=0, y index=3] {length.txt};
    \addplot [red,solid,mark=diamond*] table [x index=0, y index=4] {length.txt};
    \addplot [black,solid,mark=diamond*] table [x index=0, y index=5] {length.txt};

%                   \addplot [black,solid,mark=+] table [x index=0, y index=1] {length.txt};
    %%\addplot [blue,dashed,mark=square*] table [x index=0, y index=2] {axis/model_select.txt};
%    \addplot [green,dashed,mark=otimes*] table [x index=0, y index=3] {axis/model_select.txt};
%    %\addplot [cyan,dashed,mark=diamond*] table [x index=0, y index=4] {axis/model_select.txt};
%   \addplot [pink,densely dashed,mark=triangle*] table [x index=0, y index=5] {axis/model_select.txt};
%    %\addplot [black,solid,mark=+] table [x index=0, y index=6] {axis/model_select.txt};
%    \addplot [yellow,solid,mark=*] table [x index=0, y index=7] {axis/model_select.txt};
%    %\addplot [purple,solid,mark=square*] table [x index=0, y index=8] {axis/model_select.txt};
%    %\addplot [red,dotted,mark=*] table [x index=0, y index=1] {axis/LayerWise_2_dev.txt};
%    %\addplot [blue,dotted,mark=*] table [x index=0, y index=1] {axis/LayerWise_3_dev.txt};
%    %\addplot [green,dotted,mark=*] table [x index=0, y index=1] {axis/LayerWise_4_dev.txt};
%    %\addplot [pink,dotted,mark=*] table [x index=0, y index=1] {axis/LayerWise_5_dev.txt};
%    %\addplot table [x=k, y=baseline] {rerank-dev.txt};
    \end{axis}
\end{tikzpicture}
  \caption{Study of model sensitivity on document length on IMDB. All test samples are sorted by their length and divided into 10 parts. Left most dot means classification accuracy on the shortest 10\% samples. X-axis is length ranking from 0\% to 100\%. }
  \label{fig:len}
\end{figure}
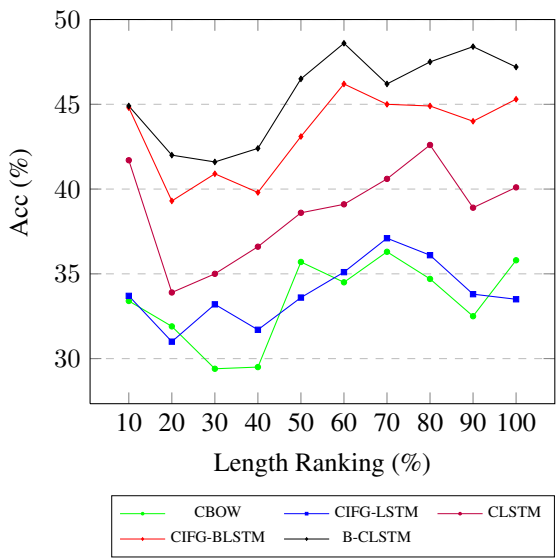

\subsection{Rate of Convergence}
We compare the convergence rates of our models, including CIFG-LSTM, CIFG-BLSTM and B-CLSTM, and the baseline models (LSTM and BLSTM). We configure the hyper-parameter to make sure every competing model has approximately the same numbers of parameters, and various models have shown different convergence rates in Figure \ref{fig:conv}.
In terms of convergence rate, B-CLSTM beats other competing models. The reason why B-CLSTM converges faster is that the splitting memory groups can be seen as a better initialization and constraints during the training process.

\subsection{Effectiveness on Grouping Memory}

For the proposed model, the number of memory groups is a highlight. In Figure \ref{fig:mem}, we plot the best prediction accuracy (Y-axis) achieved in validation set with different number of memory groups on all datasets.
From the diagram, we can find that our model outperforms the baseline method. In Yelp 2013, when we split the memory into 4 groups, it achieves the best result among all tested memory group numbers.
We can observe the dropping trends when we choose more than 5 groups. %This can be explained from two views.

For fair comparisons, we set the total amount of neurons in our model to be same with vanilla LSTM. Therefore, the more groups we split, the less the neurons belongs to each group, which leads to a worse capacity than those who have sufficient neurons for each group.

%\begin{itemize*}
%    \item
%    \item Addicted forgetting means remembering nothing. ?
%\end{itemize*}

% Thus, the global representation is almost same as the randomly initialization memory.
% 2) Splitting memory into groups and applying different forgetting rate can be regarded as building hierarchical architecture softly and implicitly. Without prior separation of sentences and paragraphs, proposed model capture a bottom-up data flow, from local features and words to global and contextual representation. It can be explained why IMDB requires larger memory group number due to its average length doubles other two dataset.\\
\subsection{Sensitivity on Document Length}

We also investigate the performance of our model on IMDB when it encodes documents of different lengths. Test samples are divided into 10 groups with regard to the length.
From Figure \ref{fig:len}, we can draw several thoughtful conclusions.
\begin{enumerate*}
    \item Bidirectional models have much better performance than the counterpart models.
    \item The overall performance of B-CLSTM is better than CIFG-BLSTM. This means that our model is adaptive to both short texts and long documents. Besides, our model shows power in dealing with very long texts in comparison with CIFG-BLSTM.
    \item CBOW is slightly better than CIFG-LSTM due to LSTM forgets a large amount of information during the unidirectional propagation.
\end{enumerate*}

\section{Conclusion}
In this paper, we address the problem of effectively analyzing the sentiment of document-level texts in an RNN architecture. Similar to the memory structure of human, memory with low forgetting rate captures the global semantic features while memory with high forgetting rate captures the local semantic features.
%The long-term and short-term memories interact with each other at every step.
Empirical results on three real-world document-level review datasets show that our model outperforms state-of-the-art models by a large margin.%., while some of previous models even utilize the external user and product information.  %Following study on convergence speed, number of memory groups and sensitivity on different length scale gives more intuitive and deepen explanation on our model and related model.

For future work, we are going to design a strategy to dynamically adjust the forgetting rates for fine-grained document-level sentiment analysis.
%according to different memory contents
%
\section*{Acknowledgments}
We appreciate the constructive work from Xinchi Chen. Besides, we would like to thank the anonymous reviewers for their valuable comments. This work was partially funded by National Natural Science Foundation of China (No. 61532011 and 61672162), the National High Technology Research and Development Program of China (No. 2015AA015408).

\bibliography{related,sentiment,qiu,xu}
\bibliographystyle{emnlp2016}

\end{document}